%% file: main.tex
\documentclass[runningheads]{eccv2026/llncs}

\usepackage{eccv2026/eccv}

\usepackage{eccv2026/eccvabbrv}

\usepackage{amsmath}
\usepackage{amssymb}
\usepackage{mathtools}
\usepackage{graphicx}
\usepackage{booktabs}
\usepackage{microtype}
\usepackage{graphicx}
\usepackage{subcaption}
\usepackage{booktabs} 
\usepackage{tabularx}
\usepackage{pifont}
\usepackage{dblfloatfix}
\usepackage[table]{xcolor}
\usepackage{pgfplots}
\usepackage{pgfplotstable}
\usetikzlibrary{intersections}
\usepgfplotslibrary{fillbetween}
\usepackage{placeins}
\definecolor{lightgray}{gray}{0.88}

\definecolor{cyan}{RGB}{70, 240, 240} 
\definecolor{pink}{RGB}{250, 190, 212} 
\definecolor{olive}{RGB}{128, 128, 0} 

\newcommand{\ours}{EPOS-VLM} 
\newcommand{\ourslong}{Embodied Persistent Object Semantics}

\newcommand{\simca}{Galliena et al.~\cite{Galliena_2025_ICCV}}
\newcommand{\pseudocaptioner}{3D-CPS}
\newcommand{\pseudocaptionerlong}{3D-Aware Consistent Pseudo-Captioner}

\usepackage[showdeletions]{color-edits} 
\addauthor[Ste]{ste}{red}
\addauthor[Tommaso G.]{tommasoG}{blue}
\addauthor[Tommaso A.]{tommasoA}{green}
\addauthor[lorenzo]{lorenzo}{cyan}

\usepackage[accsupp]{axessibility}  

\usepackage[pagebackref,breaklinks,colorlinks,citecolor=eccvblue]{hyperref}

\usepackage{orcidlink}

\usepackage[capitalize,noabbrev]{cleveref}

\usepackage[textsize=tiny]{todonotes}

\newcommand{\bbox}{\protect\raisebox{1pt}{\protect\tikz \protect\draw[black,fill=black] (1,1) circle (0.5ex);}}
\newcommand{\wbox}{\protect\raisebox{1pt}{\protect\tikz \protect\draw[black,fill=white] (1,1) circle (0.5ex);}}

\pgfplotsset{compat=1.18}

\begin{document}

\title{Memory-Augmented Vision–Language Agents for Persistent and Semantically Consistent Object Captioning}

\titlerunning{Memory-Augmented Vision–Language Agents} 

\author{Tommaso Galliena\inst{1, 2} \and
Stefano Rosa\inst{1} \and Tommaso Apicella \inst{1} \and Pietro Morerio \inst{1} \and Alessio Del Bue \inst{1} \and Lorenzo Natale \inst{1} }

\authorrunning{T.Galliena et al.}

\institute{Italian Institute of Technology, Genoa, Italy \and
University of Genoa, Genoa, Italy}

\maketitle

\begin{figure*}[t]
    \centering
    \includegraphics[width=\textwidth]{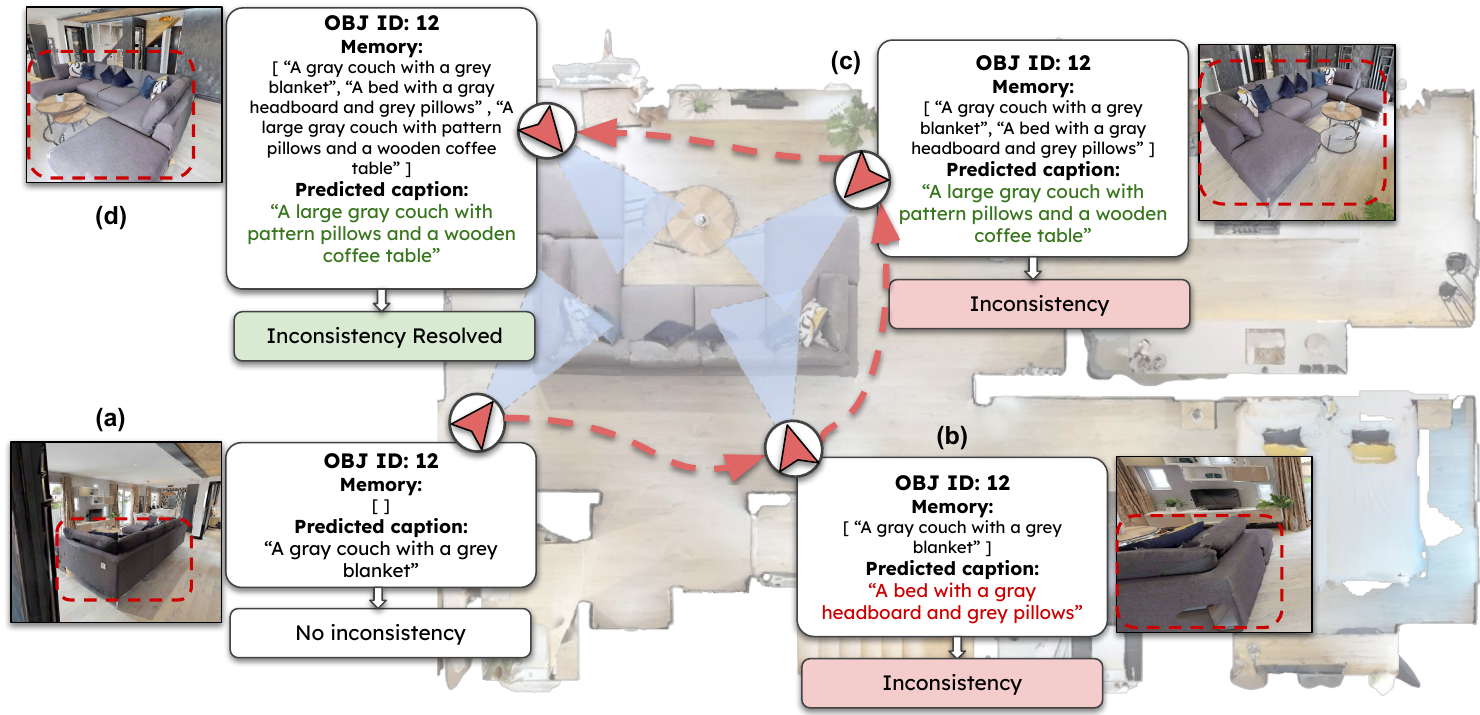}
    \caption{Memory-driven multi-view exploration progressively resolves ambiguous object captions into a consistent object-level description. (a) the agent predicts a caption for the observed object; (b) a different caption is predicted from a different viewpoint; (c) a consistent caption is predicted based on the episodic object memory; (d) the predicted caption for the object remains consistent.}
    \label{fig:scenario}
\end{figure*}

\begin{abstract}
Vision–Language Models (VLMs) often yield inconsistent descriptions of the same object across viewpoints, hindering the ability of embodied agents to construct consistent semantic representations over time. Previous methods resolved inconsistencies using offline multi-view aggregation or multi-stage pipelines that decouple exploration, data association, and caption learning, with limited capacity to reason over previously observed objects. In this paper, we introduce a unified, memory-augmented Vision–Language agent that simultaneously handles data association, object captioning, and exploration policy within a single autoregressive framework. The model processes the current RGB observation, a top-down explored map, and an object-level episodic memory serialized into object-level tokens, ensuring persistent object identity and semantic consistency across extended sequences. To train the model in a self-supervised manner, we collect a dataset in photorealistic 3D environments using a disagreement-based policy and a pseudo-captioning model that enforces consistency across multi-view caption histories. 
Extensive evaluation on a manually annotated object-level test set, demonstrate improvements of up to +11.86\% in standard captioning scores and +7.39\% in caption self-similarity over baseline models, while enabling scalable performance through a compact scene representation. 
Code, model weights, and data are available at \url{https://hsp-iit.github.io/epos-vlm/}.
\end{abstract}

\section{Introduction}
Humans actively construct robust visual understanding through dynamic, embodied exploration. From early infancy, perception is tightly coupled with action: learners build stable object representations by moving, approaching, rotating objects, and revisiting them over time~\cite{smith2005development, pexman2019role, wellsby2014developing}. 
Therefore, semantic consistency emerges from memory and repeated observations across diverse viewpoints rather than being an inherent property of independent visual inputs~\cite{smith2005development}.

In contrast, modern vision-language models (VLMs) are predominantly trained and evaluated on static datasets. Although methods such as BLIP-2~\cite{li2023blip}, Qwen-VL~\cite{yang2025qwen3technicalreport}, and LLaVA~\cite{liu2023visualinstructiontuning} achieve state-of-the-art captioning performance on single images, they exhibit severe limitations when deployed in embodied environments, including semantic inconsistency across viewpoints, lack of object permanence, and limited ability to integrate information over time.
When an agent equipped with a VLM navigates a 3D scene, changes in viewpoint, distance, and occlusion frequently cause the same physical object to be described in inconsistent or contradictory ways~\cite{Galliena_2025_ICCV}.
This semantic drift breaks object identity over time and undermines the use of language as a persistent representation for embodied reasoning, mapping, and decision-making.

Prior research addressed semantic inconsistency across viewpoints through multi-view supervision~\cite{hong20233d}, post-hoc aggregation~\cite{chan2023ic3}, or a combination of both~\cite{Galliena_2025_ICCV}, decoupling exploration, data association, captioning, and learning. In these frameworks, exploration is either confined to offline data collection or not explicitly modeled. 
Consequently, agents cannot adapt viewpoint selection based on the inconsistency of their current predictions, precluding correction of observational or aggregation errors through subsequent exploration. This decoupling fundamentally constrains the system's capability to actively acquire informative views that resolve ambiguities in object identity and attributes, particularly in novel environments.  
In contrast, we propose a unified formulation where semantic consistency is optimized end-to-end during embodied navigation. Our exploration policy directly selects viewpoints to enhance fine-grained object semantics, enabling active resolution of perceptual ambiguities during navigation (see Fig.~\ref{fig:scenario}).

We introduce \ours\ (\ourslong), a unified Vision-Language model for embodied navigation, fine-tuned end-to-end for joint data association, object-level captioning, and action prediction. 
A structured episodic memory stores per-object caption histories and 3D positions. This memory is tokenized, allowing the pretrained VLM backbone to reason over long-horizon object histories within a single autoregressive sequence. 
During training, the agent explores 3D environments, accumulating multi-view object observations. Object-level pseudo-captions, aggregated from memory histories, supervise fine-tuning of the same VLM. 
Joint learning of captioning, data association, and action selection yields policies that revisit semantically unstable objects and collect maximally informative viewpoints for ambiguity resolution.

Our contributions include: (1) a unified, memory-conditioned Vision-Language-Action model specifically designed for embodied captioning, which jointly learns data association, object-level captioning, and action prediction; (2) a structured tokenization of episodic object memory to enable end-to-end reasoning over long-horizon object histories within a pretrained VLM; and (3) an embodied captioning dataset containing navigation data and object-level pseudo-captions, along with a manually-annotated benchmark of object-level captions. 

\section{Related Work}
We contextualize our contributions within three interconnected research areas: cross-view semantic consistency in embodied captioning, memory mechanisms for object-level representation, and action-driven exploration for semantic learning.

\textbf{Cross-View Semantic Consistency in Captioning.}
While modern Vision-Language Models (VLMs) achieve strong performance on standard image captioning benchmarks
\cite{li2023blip,yu2022cocacontrastivecaptionersimagetext,alayrac2022flamingovisuallanguagemodel}, they are trained on independent image-text pairs and lack explicit mechanisms to enforce semantic consistency across observational variants (e.g., viewpoint shifts, occlusions, or scale changes). Consequently, embodied agents using off-the-shelf VLMs produce inconsistent object-level descriptions under environmental dynamics—a limitation not exposed by static image captioning evaluations~\cite{Bigazzi_2021}. 
Recent work addresses this limitation through post-hoc aggregation: ECO-score~\cite{jeong2024technical} selects captions via semantic and CLIP-based similarity, while IC3~\cite{chan2023ic3} and~\cite{Galliena_2025_ICCV} synthesize view-consistent descriptions using consensus or frequency-based in-context learning. 
Critically, these approaches only perform aggregation~\cite{chan2023ic3} or decouple aggregation from captioner training~\cite{Galliena_2025_ICCV}, treating consistency as an emergent property rather than an integrated objective.
Our work bridges this gap by unifying cross-view consistency into the captioning process via end-to-end optimization.

\textbf{Object-level Memory for Embodied Representation.}
Memory augmentation has proven effective for long-horizon embodied reasoning, with frameworks like Mem2Ego~\cite{zhang2025mem2egoempoweringvisionlanguagemodels} and MemoryVLA~\cite{shi2025memoryvlaperceptualcognitivememoryvisionlanguageaction} leveraging episodic memory to resolve partial observability in navigation and manipulation. Similarly, 3DLLM-Mem~\cite{hu20253dllmmemlongtermspatialtemporalmemory} maintains spatial-temporal memory for persistent scene understanding. However, these systems optimize memory for task execution (e.g., path planning or action prediction), not to create accurate and consistent object descriptions. 
Additionally, these methods treat memory retrieval and caption generation as sequential stages, inhibiting joint optimization of data association and description synthesis. 
In contrast, our framework embeds object-level memory directly into the VLM architecture, enabling co-optimization of cross-view aggregation, data association, and autoregressive caption generation. 
Recent work has explored structured object-level tokenization to align 3D representations with language and vision in a unified autoregressive framework~\cite{sahoo2026aligningtextimages3d}. Inspired by this object-level tokenization paradigm, our framework embeds episodic object memory directly into the captioning architecture, enabling joint optimization of cross-view aggregation, data association, and autoregressive caption generation.

\textbf{Action-Driven Semantic Learning.}
Active exploration strategies can enhance semantic consistency by strategically gathering informative observations. Systems like CaBOT~\cite{hu2023explore} and Explore-and-Explain~\cite{Bigazzi_2021} demonstrate that viewpoint selection impacts caption informativeness, while ``Where to Learn''~\cite{10877883} formalizes exploration policies for visual concept acquisition. 
Galliena et al.~\cite{Galliena_2025_ICCV} further demonstrate that caption disagreement can be used as an intrinsic signal to guide exploration and distill object-level pseudo-captions in a self-supervised setting.  
Nevertheless, existing methods decouple exploration from captioning: policies are either heuristic-based~\cite{hu2023explore} or trained independently of the captioner~\cite{10877883}, preventing online operation where evolving semantic uncertainty guides future observations.
Our approach uniquely integrates action selection, data association, and captioning into a single end-to-end framework, where uncertainty directly informs exploration to maximize cross-view consistency.

\section{Embodied Persistent Object Semantics Visual Language Model (\ours)}
\label{sec:method}

We propose a single memory-conditioned policy to explore an environment and create a consistent language representation for each encountered object. 

\begin{figure*}[t]
    \centering
    \includegraphics[width=\textwidth]{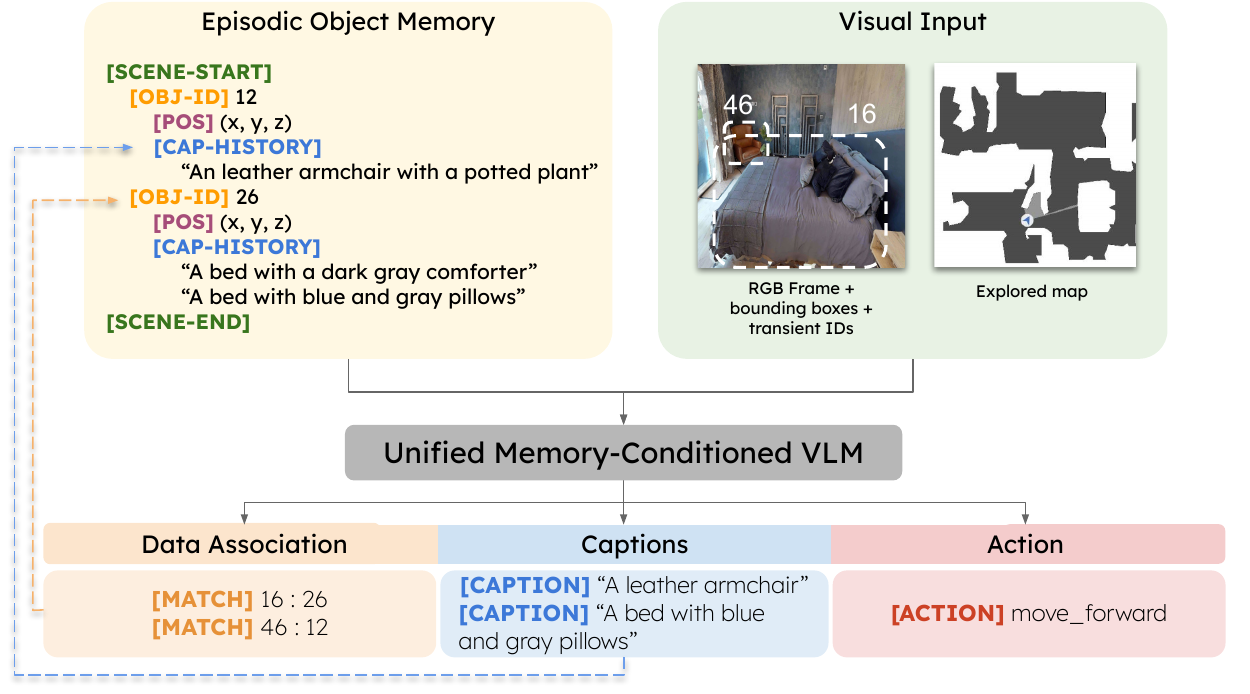}
    \caption{Overview of the proposed \ours\ model.
Structured episodic memory, RGB observations (with detected object bounding boxes and IDs overlayed), and top-down maps are jointly encoded and processed by a Vision–Language transformer, which autoregressively outputs object association, object-level captions, and actions.}
    \label{fig:vla-model}
\end{figure*}

An embodied agent navigates an environment containing a set of objects $\mathcal{O} = \{ o_i \}_{i=1}^N$ over time steps $t \in \{ 1, \dots, T \}$, processing an RGB-D observation $\mathcal{I}_t \in \mathbb{R}^{H \times W \times 4}$ from its current viewpoint and an RGB top-down explored map $\mathcal{E}_t \in \mathbb{R}^{H \times W \times 3}$ encoding visited regions.
Since objects may be observed multiple times with varying appearances, the agent maintains an 
episodic object memory $\mathcal{M}$ that assigns persistent identifiers to encountered objects and accumulates their spatial and semantic information.  
As shown in Fig.~\ref{fig:vla-model}, \ours\ conditions on $\mathcal{I}_t$, $\mathcal{E}_t$, and $\mathcal{M}_{t-1}$ to predict:  
(1) data associations linking current detections to existing object IDs or initializing new ones,  
(2) object-level captions for observed instances, and  
(3) a navigation action $a_t \in \mathcal{A}$, where $A$ is a discrete action set (e.g., \texttt{move\_forward}, \texttt{turn\_left}).  
The objective is to learn a function $\Phi_{\theta}(\mathcal{I}_t, \mathcal{E}_t, \mathcal{M}_{t-1})$ that predicts spatially consistent, viewpoint-invariant captions $\tilde{c}_o$ for objects $o$ visible in $\mathcal{I}_t$, while jointly predicting navigation actions $a_t$ that maximize object knowledge acquisition.

\subsection{Visual Observations and Object Identifiers}
\label{sec:object_proposals}
From observation $\mathcal{I}_t$, a set of object instances is extracted using an instance segmentation model, yielding a collection of detections $\{ b_t^j, p_t^j \}_{j=1}^{K_t}$, where $b_t^j \in \mathbb{R}^4$ denotes the bounding box of the $j$-th instance, $p_t^j \in \mathbb{R}^3$ its the estimated global 3D position obtained with depth projection and camera pose, and $K_t$ denotes the number of detected objects. 
To enable explicit object-level reasoning within an autoregressive vision-language model, each detected instance is assigned a transient object identifier $r_t^j \in \mathbb{N}$. Transient IDs are randomly sampled at each time step, do not correspond to persistent object identities contained in the episodic memory and
are rendered directly onto the RGB observation alongside the corresponding bounding boxes. This visual annotation allows the model to unambiguously refer to individual detections when producing data association decisions or object-level captions when multiple detections are present in the image. 
The task of linking detections across time and maintaining persistent object
identity is handled explicitly through the episodic memory mechanism described
in the following section, rather than being inferred implicitly from visual
appearance alone.

\subsection{Episodic Object Memory Representation and Tokenization}
\label{sec:memory_tokenization}

The episodic object memory provides a persistent representation of previously observed physical objects and enables long-horizon reasoning within a standard autoregressive Vision-Language model.
Formally, at time step $t$ the episodic memory is represented as a set of object entries $\mathcal{M}_t = \{ o_{\text{id}, o},\; \mathbf{p}_{t,o},\; \mathcal{C}_{t,o} \}_{o \in \mathcal{O}_t}$
where $o_{\text{id}, o}$ denotes a persistent object identifier assigned to a physical object, $\mathbf{p}_{t,o} \in \mathbb{R}^3$ is the estimated position of the object, and $\mathcal{C}_{t,o}$ stores the set of captions
previously generated for the object together with their occurrence counts,
capturing how often specific descriptions have been produced for the same object
during exploration. 

The position $\mathbf{p}_{t,o}$ is updated whenever a new observation is
associated with object $o$ using depth-based projection and camera pose, while
$\mathcal{C}_{t,o}$ is updated by either incrementing caption count for $o$ if it already contains the predicted caption or by inserting a new caption for $o$.

To condition a pretrained Vision-Language Model on episodic memory without
architectural modifications, the memory $\mathcal{M}_{t}$ 
is serialized into a
structured textual representation and injected directly into the language
context.
Each object entry is encoded as a contiguous block of text using special tokens:
\begin{equation}
\texttt{[OBJ]}\; o_{\text{id}, o}\;
\texttt{[POS]}\; (\hat{x}, \hat{y}, \hat{z})\;
\texttt{[CAPTIONS]}\; \{ (c_i, f_i) \}_{i=1}^{L},
\end{equation}
where $(\hat{x}, \hat{y}, \hat{z})$ denotes a discretized version of the object’s
global position $\mathbf{p}_{t,o}$ mapped to numeric tokens, and each pair
$(c_i, f_i)$ represents a previously generated caption $c_i$ and the number of
times $f_i$ it has been produced for that object. $L$ is the number of unique captions predicted for $o_{id}$.
All object blocks are concatenated and enclosed between the scene-level delimiters \texttt{[SCENE-START]} and \texttt{[SCENE-END]}.

\subsection{Data Association}
\label{sec:data_association}

At each time step $t$, the agent observes a set of object detections indexed by
transient identifiers $\{ r_t^j \}_{j=1}^{K_t}$ (Sec.~\ref{sec:object_proposals})
and maintains an episodic memory $\mathcal{M}_{t}$ containing entries indexed
by persistent object identifiers $\{ o_{\text{id}} \}$ (Sec.~\ref{sec:memory_tokenization}).
The goal of data association is to determine, for each current detection, whether
it corresponds to an existing memory entry or represents a previously unseen
physical object.

Data association is performed explicitly through structured \texttt{[MATCH]}
tokens generated by the autoregressive model as part of its output sequence.
Each \texttt{[MATCH]} token establishes a symbolic link between a transient object
identifier in the current RGB observation and a persistent object identifier
stored in episodic memory.
Concretely, for each transient identifier $r_t^j$, the model emits a decision of
the form
\begin{equation}
\texttt{[MATCH]}\; r_t^j \;\rightarrow\; o_{\text{id}, o}
\quad \text{or} \quad
\texttt{[MATCH]}\; r_t^j \;\rightarrow\; \texttt{[NEW\_ID]},
\end{equation}
where mapping to an existing $o_{\text{id}, o}$ indicates that the detection is
associated with a previously observed object, while the special token
\texttt{[NEW\_ID]} signals that the detection corresponds to a novel physical
object not yet present in memory.

When a detection is matched to an existing persistent identifier $o_{\text{id}, o}$,
the corresponding memory entry is updated using the current observation,
refining the object’s estimated position $\mathbf{p}_{t,o}$ and updating the
stored caption statistics $\mathcal{C}_{t,o}$.
When \texttt{[NEW\_ID]} is emitted, a new memory entry is created and initialized
from the current detection, assigning a fresh persistent identifier and storing
its initial spatial and semantic information.

Crucially, persistent object identifiers are never predicted directly from raw
visual appearance.
Instead, identity assignment emerges from joint reasoning over the current visual
evidence, spatial proximity, and the semantic information accumulated in episodic
memory.
By integrating \texttt{[MATCH]} tokens into the same autoregressive stream that
produces captions and actions, \ours\ jointly learns object linking, semantic
aggregation, and long-horizon identity maintenance within a single unified
policy.

\subsection{Training \ours}
\label{sec:policy}

\textbf{Pseudo-caption generation.} To supervise viewpoint-invariant object semantics, we aggregate multi-view observations from exploration episodes into a single object-level pseudo-caption $\tilde c_o$ per physical object $o$ using a novel 3D-grounded pseudo-captioning method, \pseudocaptionerlong\ (\pseudocaptioner), designed to produce view-consistent object descriptions. This process leverages two complementary sources: caption history and geometrically grounded visual observations, producing compact descriptions of object-intrinsic attributes while filtering spurious visual cues from occlusions or background clutter.

We store all captions generated for object $o$ during exploration in a caption-frequency dictionary $\mathcal{C}_{o}$ that estimates descriptions consistency across viewpoints: correct attributes appear frequently, while view-specific or erroneous descriptions occur sporadically. $\mathcal{C}_{o}$ thus serves as a semantic prior for viewpoint-robust linguistic attributes.

Using accumulated 3D points for object $o$, we estimate its spatial extent and select object-level image crops that maximize surface coverage of the reconstructed object. Candidate crops are ranked by distinct 3D region coverage and viewpoint diversity, prioritizing complementary views that expose defining attributes while minimizing redundancy and occlusion. 

An external vision-language model generates a pseudo-caption $\tilde{c}_o$ using (i) the cross-view caption frequency table $\mathcal{C}_o$ and (ii) the geometrically selected image crops. The model summarizes only multi-view supported attributes, explicitly omitting perspective-dependent details (e.g., relative position or camera perspective). This yields a concise pseudo-caption capturing object-intrinsic properties for self-supervised training.

\textbf{Fine-tuning.} The object-level pseudo-captions $\tilde{c}_o$ generated by \pseudocaptioner\ are then used as self-supervised targets to train \ours: for every frame in which object $o$ is observed, the model is supervised to generate $\tilde{c}_o$ (rather than a view-specific caption), enforcing a single, viewpoint-invariant description per object across the episode.

We fine-tune a pretrained vision-language model into a unified autoregressive policy \(\Phi_\theta\) that jointly performs data association, object-level captioning, and action selection.
At time step \(t\), the model conditions on the unified context  
$
X_t = [\mathcal{I}_t,\; \mathcal{E}_t,\; \mathcal{M}_{t-1}]
$. The policy autoregressively generates a structured output sequence  
\begin{equation}
Y_t = \big[ \texttt{[MATCH]}\;\Pi_t\;\; \texttt{[CAPTION]}\;C_t\;\; \texttt{[ACTION]}\;a_t \big],
\end{equation}  
with \(\Pi_t\) encoding data association (matching memory IDs or emitting \texttt{[NEW\_ID]}), \(C_t\) providing object-level captions, and \(a_t\) specifying navigation action.

Parameters \(\theta\) are optimized by minimizing the cross-entropy loss over the token sequence, i.e., maximizing the conditional likelihood of the target structured outputs \(Y_t^\star\) given the context \(X_t\). The supervision spans data association tokens, caption tokens, and action tokens within a single unified sequence.
By enforcing correct object linking, semantic consistency, and coherent actions within a single autoregressive stream, 
the model to jointly processes object identity, semantic evidence, and action outcomes, rather than treating subtasks as separable stages.

\section{Experimental Setup}
We evaluate \ours\ across five complementary tasks: (i) object-level pseudo-caption accuracy, (ii) object-level caption accuracy, (iii) cross-view semantic consistency, (iv) data association and computational scalability, and (v) impact of the action policy on semantic learning. This assessment determines whether the model improves captioning performance while maintaining persistent object semantics through embodied navigation. 

\textbf{Data Collection.} Simulations enable repeatable evaluation of long-horizon embodied behavior with access to depth, pose, and instance identifiers—essential for object-level data association and consistency metrics, which remain challenging to obtain at scale in real-world settings. We conduct experiments in Habitat~\cite{savva2019habitatplatformembodiedai} using HM3D~\cite{ramakrishnan2021habitatmatterport3ddatasethm3d} and Gibson~\cite{xia2018gibsonenvrealworldperception}. Scene-level splits follow the setup of Galliena et al.~\cite{Galliena_2025_ICCV}, but episodes are capped at 400 steps: HM3D training set consists of 128 training scenes, 27 validation scenes, and 28 test scenes. To assess out-of-domain generalization, we evaluate the model trained on HM3D on a manually annotated object-level test set from Gibson~\cite{xia2018gibsonenvrealworldperception}, comprising 32 scenes.
To collect the training supervisory signals of \ours, an heuristic-based exploration policy targets objects that exhibit high cross-view caption disagreement, and a pseudo-captioner assigns to each object instance a unique concise caption. The disagreement computation follows the methodology of Galliena et al.~\cite{Galliena_2025_ICCV}: a Mask2Former detector~\cite{cheng2022maskedattentionmasktransformeruniversal} with Swin backbone~\cite{liu2021swin} (COCO-pretrained~\cite{lin2014microsoft}) predicts instance segmentation masks and bounding boxes from input images; an image captioner (Qwen3-VL-2B~\cite{yang2025qwen3technicalreport}) predicts captions for each bounding box; a semantic point cloud created using the agent's pose and depth images embeds the segmentations and captions information; 3D object instances are segmented via connected components algorithm~\cite{cc3d}, and an object disagreement is computed as average cosine distance between all embeddings belonging to the object.
The disagreement is then down-projected onto the ground plane, forming a 2D disagreement map.
We sample candidate viewpoints around high-disagreement regions at varying distances and orientations from each object and ranked by disagreement.
Finally, training trajectories are collected while the agent navigates to each candidate viewpoint. Given that pseudo-captions are generated offline at the end of the exploration, we use a larger model (Qwen3-VL-30B~\cite{yang2025qwen3technicalreport}). We report additional details in Supp. Mat. 

\textbf{Training details.} We fine-tune \ours\ for $3$ epochs using AdamW, a learning rate of $2\times10^{-4}$, weight decay $0.01$, and a linear warmup over $3\%$ of the training steps. Training is performed on $2\times$ NVIDIA A100 GPUs with a per-device batch size of $16$ and gradient accumulation of $4$ steps, with gradient clipping at a max norm of $1.0$ for stability. We adopt LoRA for efficient fine-tuning; the LoRA configuration is reported in the Supp. Mat.

\textbf{Methods under comparison.}
We evaluate pseudo-captioning performance against IC3~\cite{chan2023ic3}, ECO-score~\cite{jeong2024technical} and LD-CPS~\cite{Galliena_2025_ICCV}. IC3 aggregates multi-view captions via LLM to capture comprehensive details;
ECO-score ranks captions using semantic similarity and CLIP consensus;
LD-CPS generates object-level pseudo-captions via LLM using agent-predicted captions and frequency.  
For object-level captioning, we benchmark against state-of-the-art Vision-Language Models: Qwen3-VL-2B~\cite{yang2025qwen3technicalreport}, BLIP-2~\cite{li2023blip}, 
InternVL-3B~\cite{wang2025internvl35advancingopensourcemultimodal}, CoCa~\cite{yu2022cocacontrastivecaptionersimagetext}, and Florence~\cite{yuan2021florencenewfoundationmodel}, and LD-CPS+CLA pipeline~\cite{Galliena_2025_ICCV} for cross-view consistency.
We compare data association performance between our episodic object memory and \simca's point-cloud based method to evaluate the trade-off between accuracy and computational overhead.
To assess whether \ours\ predicts actions that increase caption accuracy and semantic consistency, we fix the captioning architecture while varying the action policy:  
(i) Random Goal, which randomly samples navigation targets;  
(ii) Frontier Exploration~\cite{yamauchi1998frontier}, which prioritizes geometric coverage without semantic awareness;
(iii) \ours, which conditions actions on observations, explored map, and episodic memory.

\textbf{Metrics.}
We quantify caption accuracy using complementary captioning metrics~\cite{sarto2025image}: BLEU-4 (B4)~\cite{papineni2002bleu}, METEOR (M)~\cite{banerjee2005meteor}, ROUGE-L (RL)~\cite{lin2004rouge}, CIDEr 
(CI)~\cite{vedantam2015cider}, and SPICE (SP)~\cite{anderson2016spice}. We use established implementations~\cite{chen2015microsoft,reimers2019sentence}, computing per object scores and averaging across test scenes. 
We evaluate cross-view semantic consistency through SBERT~\cite{reimers-2019-sentence-bert} embedding cosine similarity (CS), where higher values indicate greater consistency, supplemented by InterQuartile Range (IQR) for dispersion analysis. To isolate episodic memory's contribution, we compare the full model against a stateless ablation (\textit{w/o memory}) that processes only current RGB observations and explored maps. For data association, we use complementary object tracking metrics~\cite{s24010229,zhang2023motiontrack,wu2021track} to evaluate object identity correctness and temporal consistency during explorations: Association Accuracy (Acc) measures the proportion of correct identity assignments; F1-Match (F1-M) and F1-New (F1-N) measure the reliability in matching observations to existing objects or assigning new IDs respectively; Identity Switches (IDSW)~\cite{wu2021track} measures the amount of matches between an object identity and other existing objects, while Track Fragmentation (Frag)~\cite{s24010229} the matches with new object identities. 

\section{Results}
\subsection{Pseudo-Captioning Evaluation} 
In Tab.~\ref{tab:pseudocaption_hm3d} we compare our pseudo-captioner \pseudocaptioner\ with previous methods. \pseudocaptioner\ consistently outperforms all baselines in pseudo-captioning performance, showing that processing visual and language information in form of episodic memory during caption aggregation yields more accurate and semantically aligned object descriptions than baselines.
Unlike IC3~\cite{chan2023ic3}, which summarizes captions assuming that all input views are correct, our approach filters out inconsistent observations, resulting in a solid improvement across all metrics. Consequently, our method gains +32.86 percentage points (p.p.) in SPICE and +20.37 p.p. in caption similarity (CS).
Compared to LD-CPS~\cite{Galliena_2025_ICCV}, which relies purely on language-based aggregation, our method leverages visual data to resolve semantic ambiguities, improving SPICE by 8.92 p.p. and CS by +5.26 p.p.
Furthermore, while ECO-score~\cite{jeong2024technical} might select the wrong description based on the object images and captions alignment, \pseudocaptioner\ processes object images and captions to generate a more comprehensive and accurate description, outperforming ECO-score by +0.62 in CI and +6.10 p.p. in CS. Based on this analysis, we supervise \ours\ fine-tuning of the object-level captions with \pseudocaptioner's output.

\input{tex/pseudocaptioner_evaluation}

\subsection{Object-level Captioning and Semantic Consistency}

\textbf{Captioning performance.}
In Tab.~\ref{tab:caption_accuracy_consistency}, we show the performance in object-level captioning evaluated on HM3D test set. \ours\ outperforms all off-the-shelf image captioners, achieving +9.76 p.p. in SP and +0.89 in CI compared to the best performing VLM (Intern-VL~\cite{wang2025internvl35advancingopensourcemultimodal}). This result highlights that despite state-of-the-art performance in image captioning, off-the-shelf VLMs are not tailored for embodied settings. \ours's performance surpasses also Galliena et al.'s multi-stage pipeline~\cite{Galliena_2025_ICCV}, with improvements of +0.65 in CI, +2.04 p.p. in SP, and +7.12 p.p. in CS. Compared to Galliena et al.'s method, we use the full image frame instead of cropping objects, adding more context to the VLM input, and we use the episodic memory containing the correspondence between object instance and the captions from different viewpoints to disambiguate challenging viewpoints.
We compare also \ours\ full model against a variant without episodic memory, which conditions only on the current observation and map.
Removing memory prevents data association and forces the model to process each time step independently.
This results in a performance degradation of -14.15 p.p. in SP and -12.60 CS compared to the full model, confirming that episodic memory is a core component of the proposed framework.

\textbf{Cross-View Semantic Consistency.}
Tab.~\ref{tab:caption_accuracy_consistency} also shows cosine similarity across all captions generated for the same object instance and averaged over the test set. \ours\ achieves the state-of-the-art performance in Mean CS, higher than the best performing VLM (BLIP-2) by +29.22 p.p., while scoring the lowest IQR.
High IQR in baselines (more than 25) reflects inconsistent object descriptions due to viewpoint changes or partial occlusions.
In contrast, \ours\ is trained to look at the same objects from different viewpoints, and progressively refine their descriptions.
\ours\ also outperforms the consistency-based method~\cite{Galliena_2025_ICCV} by +7.39 p.p. in Mean CS and -6.08 in IQR. These results demonstrate that memorizing object-level captions and object positions during navigation increases consistency of object descriptions across viewpoint changes.
The memory-ablated variant has lower scores than the full model confirming that consistency is a direct result of the proposed episodic memory.

\input{tex/caption_accuracy_consistency}

\input{tex/gibson_results}

\textbf{Generalization to unseen environments.}
Tab.~\ref{tab:caption_accuracy_consistency_gibson} compares the performance of \ours\ with the method by Galliena et al.~\cite{Galliena_2025_ICCV}, both trained on HM3D and tested on  Gibson~\cite{xia2018gibsonenvrealworldperception}, whose scenes differ substantially in layout and appearance. \ours\ achieves higher caption accuracy with +0.58 CI and +3.46 p.p. SP, and improved alignment with the reference descriptions with +7.65 p.p. compared to the multi-stage method~\cite{Galliena_2025_ICCV}.
Crucially, \ours\ also achieves higher semantic consistency in objects descriptions, with a higher mean and median caption similarity and lower dispersion, compared to \simca.
Although Gibson represents a challenging domain shift due to the different environments and lower mesh quality compared to HM3D, the performance gap remains moderate, indicating that the learned object-level memory and multi-view aggregation mechanisms transfer effectively to previously unseen environments.
These results suggest that explicitly modeling episodic memory and enforcing semantic consistency across viewpoints leads to more robust and transferable object-level representations, which are essential for deploying embodied Vision–Language agents in diverse real-world settings.

\subsection{Data Association and Scalability}

\input{tex/data_association}

Tab.~\ref{tab:data_assoc} compares the performance between \ours\ and point cloud based method used by \simca\ in associating a persistent object identifier  to a previously seen physical object or to a new one based on a visual observation. The results show that \ours\ achieves comparable results with the point cloud based method with +0.02 p.p. in Acc, +0.02 p.p. in F1-M and -0.01 p.p. in F1-N. Despite similar performance, we represent objects using the estimated 3D position and track their change during the exploration (sparse representation), rather than a high-resolution point cloud (dense representation) embedding also the visual and language embeddings. To analyse the advantages of our memory representation, we compare the inference time and memory usage at test time for \ours\ and the point cloud baseline (see Fig.~\ref{fig:timing-comp}). The inference time of the point cloud baseline approximately exhibits a linear inference time after 100 steps.
In contrast, our method maintains a near-constant inference time of $\simeq$ 0.7s per step, attributable to leveraging a fixed-capacity tokenized memory. Similarly, the memory footprint of \ours\ remains below 10 kB, with the point-cloud baseline growing around 1 GB over the course of an episode.
Together, these results suggest that data association for object captioning does not require a dense, high-resolution object representation of the environment. Instead, our textual episodic memory containing a sparse object representation, combined with a model capable of reasoning over spatial and semantic context, achieves comparable performance while drastically reducing computational overhead and memory usage.

\input{tex/inference_time}

\subsection{Action Policy Evaluation}
\label{sec:action_eval}

To quantitatively assess whether \ours\ generates trajectories that enhance semantic consistency, we compare \ours\ exploration policy with frontier exploration~\cite{yamauchi1998frontier} and random goal exploration.
Results in Tab.~\ref{tab:policy_evaluation_consistency} show that \ours\ consistently achieves significant improvements across all metrics. Specifically, \ours\ outperforms Frontier Exploration by +0.39 p.p. in CI and +4.38 p.p. in SP. 
\ours\ also yields an improvement in cross-view caption consistency, with gains of +9.22 p.p. in Mean CS and -6.14 in IQR.  
These improvements stem from \ours\ different action selection: Frontier Exploration prioritizes geometric coverage without explicit mechanisms for resolving semantic 
ambiguity, whereas \ours\ revisits objects exhibiting inconsistent captions. 
Overall, our analysis confirms that semantic consistency is not merely a function of observation aggregation and memory retention, but critically depends on action policies that increase semantic consistency. 

\input{tex/policy_evaluation_consistency}

\subsection{Qualitative results}
\label{sec:qualitative}

\begin{figure}[t!]
    \centering
    \includegraphics[width=\linewidth]{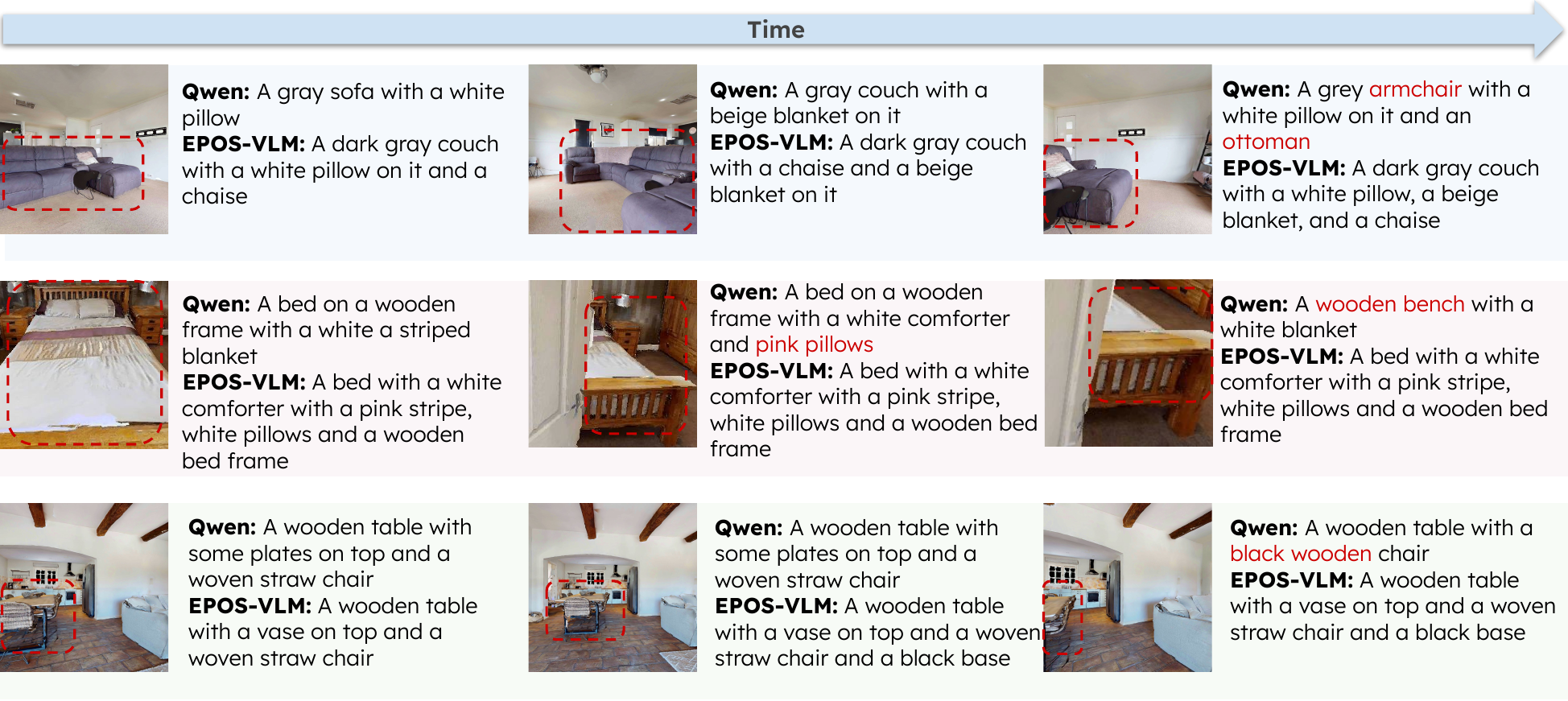}
    \caption{Comparison of object captions predicted by a VLM baseline (Qwen3-VL) and our method along successive viewpoints of an exploration. Mistakes highlighted in red.}
    \label{fig:qualitative_results}
\end{figure}

Fig.~\ref{fig:qualitative_results} compares qualitatively object captions predicted by Qwen3-VL and \ours\ across different views during exploration. 
Qwen3-VL predicts independent captions for each object crop, resulting in semantic inconsistency e.g., alternating ``sofa'', ``couch'' and ``armchair'' for the same object. 
In contrast, our model leverages explicit memory and data association to maintain persistent object identity across sequential views and to iteratively refine descriptions while preserving core attributes, yielding compact, object-level captions resilient to viewpoint variation, occlusion, and scale changes. More qualitative results in Supp. Mat. 

\section{Conclusion}  
This work introduced \ours, a memory-augmented Vision-Language-Action model for embodied captioning, addressing caption consistency of the same objects. \ours\ integrates object-level episodic memory, object descriptions association, and action selection in a single autoregressive architecture. This enables the model to process the history of the objects locations and descriptions across time and viewpoints. 
Extensive experiments demonstrate significant improvements in object-level caption accuracy, cross-view semantic consistency, long-horizon data association, and computational scalability compared to prior methods. This supports the idea that long-horizon semantic reasoning in embodied agents requires persistent memory as opposed to transient visual representation.

\textbf{Limitations and future work.} \ours\ uses an external instance segmentation model that can predict inaccurate object masks causing mistakes in data association and captioning; the experimental validation focused on analysing the performance in photorealistic and static environments assuming ideal sensors. Future work will integrate the segmentation phase inside the model, address sensor noise and real-time constraints during real-world deployment, and explore the capabilities of our memory in dynamic scenes, where new objects may appear. 

\section{Acknowledgments}
This project received funding from the European Union's Horizon research and innovation programme G.A. n. 101070227 (CONVINCE)

\bibliography{main}
\bibliographystyle{eccv2026/splncs04}

\input{tex/appendix}


\end{document}

%% file: tex/pseudocaptioner_evaluation.tex
\begin{table}[t]
\centering
\caption{Pseudo-captioning performance on manually annotated HM3D~\cite{ramakrishnan2021habitatmatterport3ddatasethm3d} test set.
}
\label{tab:pseudocaption_hm3d}
\scriptsize
\begin{tabular}{l cc cccccc}
\toprule
\textbf{Pseudo-captioner} & \textbf{LAN} & \textbf{RGB} & \textbf{B4} & \textbf{M} & \textbf{RL} & \textbf{CI} & \textbf{SP} & \textbf{CS} \\
\midrule
IC3~\cite{chan2023ic3} & \bbox & \wbox & 
1.36 & 13.56 & 15.17 & 0.03 & 11.87 & 57.89 \\
LD-CPS~\cite{Galliena_2025_ICCV} & \bbox & \wbox & 
21.02 & 26.86 & 53.10 & 1.36 & 35.81 & 73.00 \\
ECO-score~\cite{jeong2024technical} & \bbox & \bbox & 
18.36 & 23.27 & 47.32 & 1.30 & 34.98 & 72.16 \\
\rowcolor{lightgray}
\textbf{\pseudocaptioner} & \bbox & \bbox & 
\textbf{28.54} & \textbf{31.97} & \textbf{61.13} & \textbf{1.92} & \textbf{44.73} & \textbf{78.26} \\
\bottomrule
\addlinespace[\belowrulesep]
\multicolumn{9}{l}{\parbox{0.7\linewidth}{\scriptsize{
KEYS -- $LAN$: language, $RGB$: image, $B4$: BLEU, $M$: METEOR, $RL$: ROUGE-L, $CI$: CIDEr, $SP$: SPICE, $CS$: cosine similarity between SBERT embedding of prediction and annotation.}}}
\end{tabular}
\end{table}

%% file: tex/caption_accuracy_consistency.tex
\begin{table}[t]
\setlength{\tabcolsep}{2pt}
\centering
\scriptsize
\caption{Caption accuracy ($B4-CS$) and caption consistency across multiple views of the same object (Mean $CS-IQR$) on HM3D~\cite{ramakrishnan2021habitatmatterport3ddatasethm3d} manually annotated test set.}
\label{tab:caption_accuracy_consistency}
\begin{tabular}{l|cccccc|ccc}
\toprule
 & \multicolumn{6}{c}{Accuracy} & \multicolumn{3}{c}{Consistency} \\
\textbf{Model} & \textbf{B4} & \textbf{M} & \textbf{RL} & \textbf{CI} & \textbf{SP} & \textbf{CS} & \textbf{Mean CS} $\uparrow$ & \textbf{Median CS} $\uparrow$ & \textbf{IQR}
$\downarrow$ \\
\midrule
Qwen3-VL~\cite{yang2025qwen3technicalreport} & 16.00  & 15.97 & 41.15 & 0.84 & 29.88 & 60.01 & 59.65 & 60.01 & 29.12 \\
BLIP-2~\cite{li2023blip} & 10.99 & 14.51 & 39.99 & 0.57 & 26.16 & 56.88 & 60.15 & 64.12 & 25.29 \\
Intern-VL~\cite{wang2025internvl35advancingopensourcemultimodal} & 17.21 & 16.87 & 46.02 & 0.98 & 30.02 & 61.18 & 55.19 & 58.76 & 27.34 \\
CoCa~\cite{yu2022cocacontrastivecaptionersimagetext}  & 15.23 & 16.13 & 43.33 & 0.83 & 29.95 & 57.47 & 57.87 & 57.26 & 29.01 \\
Florence~\cite{yuan2021florencenewfoundationmodel}  & 14.19 & 17.26 & 44.03 & 0.78 & 31.45 & 59.12 & 52.56 & 55.00 & 26.98 \\
\simca & 19.02 & 23.04  & 48.02 & 1.12 & 39.78 & 70.00 & 81.98 & 83.07 & 8.87 \\
\rowcolor{lightgray} 
\textbf{\ours} & \textbf{25.86} & \textbf{27.89} & \textbf{59.88} & \textbf{1.87} & \textbf{41.82} & \textbf{77.12} & \textbf{89.37}  & \textbf{90.32} & \textbf{2.79} \\
\rowcolor{lightgray}
w/o memory & 16.65 & 18.01 & 45.34 & 0.85 & 27.67 & 64.62 & 52.12 & 52.84 & 32.01 \\
\bottomrule
\addlinespace[\belowrulesep]
\multicolumn{10}{l}{\parbox{\linewidth}{\scriptsize{
KEYS – $B4$: BLEU, $M$: METEOR, $RL$: ROUGE-L, $CI$: CIDEr, $SP$: SPICE, $CS$: cosine similarity between SBERT embedding of prediction and annotation, Mean CS / Median CS:
cosine similarity between SBERT embeddings of captions for the same object (higher, better), IQR: interquartile range of cosine similarity (lower, better).
}}}
\end{tabular}
\end{table}

%% file: tex/gibson_results.tex
\begin{table}[t]
\setlength{\tabcolsep}{2pt}
\centering
\scriptsize
\caption{Caption accuracy ($B4$--$CS$) and caption consistency across multiple views of the same object (Mean $CS$--IQR) on Gibson~\cite{xia2018gibsonenvrealworldperception} manually annotated test set. Methods are trained on HM3D~\cite{ramakrishnan2021habitatmatterport3ddatasethm3d}.}
\label{tab:caption_accuracy_consistency_gibson}
\begin{tabular}{l|cccccc|ccc}
\toprule
 & \multicolumn{6}{c}{Accuracy} & \multicolumn{3}{c}{Consistency} \\
\textbf{Model} 
& \textbf{B4} & \textbf{M} & \textbf{RL} & \textbf{CI} & \textbf{SP} & \textbf{CS} 
& \textbf{Mean CS} $\uparrow$ & \textbf{Median CS} $\uparrow$ & \textbf{IQR} $\downarrow$ \\
\midrule
\simca
& 17.03 & 19.65 & 46.88 & 1.05 & 35.58 & 66.03 & 78.18 & 80.10 & 11.92 \\
\rowcolor{lightgray}
\textbf{\ours} 
& \textbf{19.88} & \textbf{22.94} & \textbf{53.87} & \textbf{1.63} & \textbf{39.14} & \textbf{75.86} & \textbf{85.83} & \textbf{85.94} & \textbf{4.17} \\
\bottomrule
\addlinespace[\belowrulesep]
\multicolumn{10}{l}{\parbox{\linewidth}{\scriptsize{
KEYS – $B4$: BLEU-4, $M$: METEOR, $RL$: ROUGE-L, $CI$: CIDEr, $SP$: SPICE, $CS$: cosine similarity between SBERT embedding of prediction and annotation.
Mean CS / Median CS: cosine similarity between SBERT embeddings of captions for the same object (higher, better), IQR: interquartile range of cosine similarity (lower, better).
}}}
\end{tabular}
\end{table}

%% file: tex/data_association.tex
\begin{table}[t]
\centering
\scriptsize
\caption{Data association performance over exploration trajectories on HM3D~\cite{ramakrishnan2021habitatmatterport3ddatasethm3d} test set.}
\label{tab:data_assoc}
\setlength{\tabcolsep}{4pt}
\begin{tabular}{lccccc}
\toprule
\textbf{Method} &
\textbf{Acc} $\uparrow$ &
\textbf{F1-M} $\uparrow$ &
\textbf{F1-N} $\uparrow$ &
\textbf{IDSW} $\downarrow$ &
\textbf{Frag} $\downarrow$ \\
\midrule
\simca &
0.91 &
0.95 &
\textbf{0.90} &
0.11 &
0.08 \\
\rowcolor{lightgray} 
\textbf{\ours} &
\textbf{0.93 } &
\textbf{0.97 } &
0.89 &
\textbf{0.09} &
\textbf{0.06} \\
\bottomrule
\addlinespace[\belowrulesep]
\multicolumn{6}{l}{\parbox{0.68\linewidth}{\scriptsize{
KEYS -- $Acc$: matching accuracy, $F1-M$: F1-score for matching already observed objects, $F1-N$: F1-score for detecting newly observed objects (NEW), $IDSW$: ID switches, $Frag$: track fragmentation}}}
\end{tabular}
\end{table}

%% file: tex/inference_time.tex
\pgfplotstableread{data/inference_time_pc.txt}\timepc
\pgfplotstableread{data/inference_time_ours.txt}\timeours
\pgfplotstableread{data/inference_mem_pc.txt}\mempc
\pgfplotstableread{data/inference_mem_ours.txt}\memours

\begin{figure}[t!]
    \centering
    \begin{tikzpicture}
    \begin{axis}[
    width=0.48\linewidth,
    xmin=-10, xmax=410,
    xtick={0, 100, 200, 300, 400},
    xlabel={Step number},
    height=0.35\linewidth,
    ymin=0, ymax=2.5,
    ytick={0,0.5,1.0,1.5,2.0,2.5},
    ylabel={Time [s]},
    label style={font=\footnotesize},
    tick label style={font=\footnotesize},
    ]
    \addplot [name path=lower, fill=none, draw=none] table [x=X, y expr=\thisrow{Y} -\thisrow{STD}] {\timepc};
    \addplot [name path=upper, fill=none, draw=none] table [
    x=X, y expr=\thisrow{Y} + \thisrow{STD}]{\timepc};
    \addplot[magenta!10] fill between[of=lower and upper];
    \addplot[draw=magenta, line width=0.5pt] table [x=X, y=Y] {\timepc};
    \addplot [name path=lower, fill=none, draw=none] table [x=X, y expr=\thisrow{Y} -\thisrow{STD}] {\timeours};
    \addplot [name path=upper, fill=none, draw=none] table [
    x=X, y expr=\thisrow{Y} + \thisrow{STD}]{\timeours};
    \addplot[olive!30] fill between[of=lower and upper];
    \addplot[draw=olive, line width=0.5pt] table [x=X, y=Y] {\timeours};
    \end{axis}
    \end{tikzpicture}
    \hfill
    \begin{tikzpicture}
    \begin{axis}[
    width=0.48\linewidth,
    xmin=-10, xmax=410,
    xtick={0, 100, 200, 300, 400},
    xlabel={Step number},
    height=0.35\linewidth,
    ymode=log,
    ytick={0.000000001, 0.000001, 0.001, 1}, 
    ylabel={Memory [GB]},
    label style={font=\footnotesize},
    tick label style={font=\footnotesize},
    ]

    \addplot [name path=lower, fill=none, draw=none] table [x=X, y expr=\thisrow{Y} -\thisrow{STD}] {\mempc};
    \addplot [name path=upper, fill=none, draw=none] table [
    x=X, y expr=\thisrow{Y} + \thisrow{STD}]{\mempc};
    \addplot[magenta!10] fill between[of=lower and upper];
    \addplot[draw=magenta] table [x=X, y=Y] {\mempc};

    \addplot [name path=lower, fill=none, draw=none] table [x=X, y expr=\thisrow{Y} -\thisrow{STD}] {\memours};
    \addplot [name path=upper, fill=none, draw=none] table [
    x=X, y expr=\thisrow{Y} + \thisrow{STD}]{\memours};
    \addplot[olive!30] fill between[of=lower and upper];
    \addplot[draw=olive, line width=0.5pt] table [x=X, y=Y] {\memours};
    \end{axis}
    \end{tikzpicture}
   \caption{Mean and standard deviation of inference time (left) and memory occupancy (right) averaged over HM3D~\cite{ramakrishnan2021habitatmatterport3ddatasethm3d} test episodes for \ours\ and the dense point cloud baseline~\cite{Galliena_2025_ICCV}. 
   KEY: \protect\raisebox{2pt}{\protect\tikz \protect\draw[magenta,line width=3] (0,0) -- (0.3,0);}~Point cloud, 
    \protect\raisebox{2pt}{\protect\tikz \protect\draw[olive,line width=3] (0,0) -- (0.3,0);}~Ours} 
\label{fig:timing-comp}
\end{figure}

%% file: tex/policy_evaluation_consistency.tex
\begin{table}[t]
\centering
\scriptsize
\caption{Caption accuracy ($B4$--$CS$) and caption consistency across multiple views of the same object (Mean $CS$--IQR) varying exploration policy on HM3D~\cite{ramakrishnan2021habitatmatterport3ddatasethm3d} manually annotated test set.}
\label{tab:policy_evaluation_consistency}
\begin{tabular}{l|cccccc|ccc}
\toprule
 & \multicolumn{6}{c}{Accuracy} & \multicolumn{3}{c}{Consistency} \\
\textbf{Policy} & \textbf{B4} & \textbf{M} & \textbf{RL} & \textbf{CI} & \textbf{SP} & \textbf{CS} & \textbf{Mean CS} $\uparrow$ & \textbf{Median CS} $\uparrow$ & \textbf{IQR}
$\downarrow$ \\
\midrule
Random Goal & 21.98 & 24.12 & 52.78 & 1.38 & 33.88 & 74.02 & 80.33 & 80.65 & 9.03 \\
Frontier Exploration~\cite{yamauchi1998frontier} & 22.87 & 25.01 & 53.67 & 1.50 & 35.36 & 74.89 & 80.21 & 81.03 & 8.45 \\
\rowcolor{lightgray} 
\textbf{\ours} & \textbf{25.86} & \textbf{27.89} & \textbf{59.88} & \textbf{1.87} & \textbf{41.82} & \textbf{77.12} & \textbf{89.37} & \textbf{90.32} & \textbf{2.79} \\
\bottomrule
\addlinespace[\belowrulesep]
\multicolumn{10}{l}{\parbox{\linewidth}{\scriptsize{
KEYS -- $B4$: BLEU-4, $M$: METEOR, $RL$: ROUGE-L, $CI$: CIDEr, $SP$: SPICE, $CS$: cosine similarity between SBERT embeddings of prediction and annotation. Mean $CS$ and Median
$CS$ measure cross-view consistency (higher is better), $IQR$ measures variability (lower is better).
}}}
\end{tabular}
\end{table}

%% file: tex/appendix.tex
\newpage
\appendix

\section{\ours\ details}
\subsection{Architecture}
In all our experiments, we use a pretrained Qwen3-VL-2B~\cite{yang2025qwen3technicalreport} model as the basis for \ours.
The vocabulary is extended with the addition of the following non-learnable special tokens: \texttt{[SCENE-START]}, \texttt{[SCENE-END]}, \texttt{[OBJ-ID]}, \texttt{[POS]}, \texttt{[CAP-HISTORY]}, \texttt{[CAPTION]}, \texttt{[ACTION]}, \texttt{[MATCH]}. These are processed by the standard tokenizer and are used by the model to have a structured representation of the episodic object memory introduced in Sec. 3.2.

\subsection{Prompt Formatting}
\label{sec:supp_prompt}

This section details how the visual and textual input modalities are constructed and serialized before being passed to the Unified VLM at each timestep.

\textbf{Combined Visual Input.}
The RGB-D observation $\mathcal{I}_t$ and the explored map $\mathcal{E}_t$ are resized to dimensions $224 \times 224 \times 3$, horizontally stacked to form an input image. 
The RGB image is augmented with object-level annotations. Bounding boxes detected by the object detector are drawn on the RGB image.
Each bounding box is also labeled on the image with a random transient ID $r_t^{j \in K_t}$, where $K_t$ is the number of detected objects in the image.
The explored map shows the areas of the map that have been explored up to time step $t$, as well as the agent current position and field of view.

\textbf{Textual Prompt.}
The textual input includes a prompt with instructions followed by the serialized episodic memory $\mathcal{M}_{t-1}$. 
The episodic object memory is formatted as a sequence of object data blocks delimited by special tokens and injected into the language context as described in Sec. 3.2.

This representation provides the model with a compact, object-centric summary of past observations, including spatial location and accumulated semantic evidence.

The following is an example of a textual input:

\noindent\textbf{Prompt example.}
\small
\begin{verbatim}
[TASK-START]
Your task is object linking and action prediction.

You are given:
 - A MEMORY of previously seen objects, each with a fixed OBJ-ID.
 - A FRAME with current objects, each having a temporary OBJ-ID
   (a random ID drawn over the image).

For each object in the FRAME, decide whether it corresponds
to one MEMORY object.
If it matches, output:
  [MATCH] <frame_random_id> : <memory_obj_id>
If it is a new object, output:
  [MATCH] <frame_random_id> : NEW_ID

After matching all objects, predict the action to take:
  [ACTION] [ACTION]
Available actions are:
  move_forward, stop, turn_left, turn_right

Below are the FRAME objects and their random IDs:
  37, 16

[SCENE-START]
[OBJ-ID] 11
[CAPTION-HISTORY]
  2: "a bed with a pink and blue polka dot sheet."
[POSITION] [-7.09, 1.67, 3.79]

[OBJ-ID] 12
[CAPTION-HISTORY]
  1: "a black leather couch with a white wall in the background."
  1: "a black leather chair in a room with a white wall."
  1: "a black leather couch with a white pillow on it."
  1: "a black leather couch with a pillow on it."
  3: "a black leather couch in a living room."
[POSITION] [-2.91, 1.56, 2.59]
[SCENE-END]
\end{verbatim}
\normalsize

\subsection{Output Parsing.}
The model generates a structured autoregressive sequence containing
data association decisions, object captions, and the navigation action.
The sequence is parsed by identifying the special tokens
\texttt{[MATCH]}, \texttt{[CAPTION]}, and \texttt{[ACTION]} and extracting the
corresponding arguments.

The \texttt{[MATCH]} token encodes the data association between objects
detected in the current frame and objects stored in episodic memory.
Specifically, it follows the format
\[
\texttt{[MATCH] } r_t^j : o_{id,o},
\]
where \(r_t^j\) is the transient random identifier assigned to the \(j\)-th
detection in the current frame at timestep \(t\), and \(o_{id,o}\) is the
persistent object identifier stored in episodic memory. If the detection
corresponds to a previously unseen object, the model emits
\texttt{NEW\_ID} instead of a memory identifier.

For example, consider the following generated output:
\begin{verbatim}
[MATCH] 37 : 12
[MATCH] 16 : NEW_ID
[CAPTION] "a black leather couch with a white pillow"
[CAPTION] "a small wooden side table"
[ACTION] move_forward
\end{verbatim}

The parsing procedure extracts:
\begin{itemize}
\item Data associations: \(37 \rightarrow 12\), \(16 \rightarrow \texttt{NEW\_ID}\)
\item Object captions predicted for the current detections
\item Navigation action: \texttt{move\_forward}
\end{itemize}

The data association updates the episodic memory by linking detections
to existing object entries or creating new ones when \texttt{NEW\_ID} is
emitted. Predicted captions are added to the caption history of the
corresponding object, and the predicted action is executed by the
navigation controller.

\section{Data Collection Details}
\label{sec:supp_data_collection}
This section provides a detailed description of the trajectory collection
procedure used to build the training dataset.

\textbf{Object detection and caption prediction.}
At each time step, the RGB observation is processed using Mask2Former to generate instance segmentation masks and corresponding bounding boxes. 
Each detected instance is assigned a transient, randomly sampled identifier, reinitialized independently per frame and explicitly decoupled from persistent object identities $o_{id,o}$. This design precludes reliance on temporal identity continuity, thereby mitigating the risk of shortcut learning via ID memorization during training.

For each bounding box, an object-centric crop is extracted and processed by a pretrained image captioning model (Qwen3-VL-2B~\cite{yang2025qwen3technicalreport}) to produce a descriptive caption. The caption is subsequently encoded into a fixed-dimensional text embedding using a SentenceTransformer (SBERT) \cite{reimers2019sentence} model. This embedding serves as the input for the following disagreement map construction module.

\textbf{Disagreement map construction.}
As in \cite{Galliena_2025_ICCV}, disagreement is defined as semantic inconsistency between all captions predicted for the same object.
Following \cite{Galliena_2025_ICCV}, the agent maintains a spatial disagreement map over the navigable area of the environment, defined in the top-down map frame. Intuitively, each map cell
accumulates a score reflecting semantic instability associated with observations projected into that region over time. In practice, this is implemented by aggregating caption embedding statistics over the reconstructed point cloud and projecting the resulting disagreement values into the map coordinate system using Habitat’s world-to-map transformation.

\textbf{Target object selection.}
The disagreement map is normalized and thresholded to isolate high-disagreement areas.
2D connected components is used to cluster the thresholded disagreement into target regions; regions with an area below a threshold $A_{\min}$ are discarded since they might be artifacts or represent only part of an object.
Each target region is represented by a 2D centroid.
Target regions are then ranked by disagreement and processed in descending order.

\textbf{Candidate viewpoint generation.}
For each cluster, we generate candidate viewpoints around its boundary. Candidate viewpoints are selected at $N_r$ radial distances in the set $\mathcal{R}$ from the target object and oriented towards towards the centroid.
Each candidate viewpoint is filtered using a navigability check with a safety margin to ensure the camera is not placed too close to obstacles. 
Finally, a random number of viewpoints between $N_{\min}$ and $N_{\max}$ is selected from the candidates.

\textbf{Viewpoints selection.}
Viewpoints are ranked using a heuristic that combines the target object's disagreement, weighted by a coefficient $\alpha$, and the travel distance cost from the agent's position to the viewpoint, weighted by $1-\alpha$.
Viewpoints are then sent as goals to the agent from the highest ranking to the lowest. $\alpha$ was empirically chosen to balance the agent visiting high disagreement viewpoints first while also prioritizing nearby objects.

\textbf{Navigation.}
The agent navigates to each viewpoint using the shortest path planner provided by Habitat sim~\cite{savva2019habitatplatformembodiedai}.
At each timestep the disagreement statistics are updated for all the objects present in the current field of view of the agent. 

\textbf{Failure recovery.}
A failure recovery policy detects when the agent becomes stuck while navigating toward a goal, ensuring that no exploration steps are lost. 
This condition is identified by monitoring the displacement of the agent over a sliding window of $\tau_s$ steps. 
If the displacement remains below a threshold $\epsilon_p$, the agent is considered stuck.
When this condition is detected, a recovery behaviour is triggered by sampling candidate target positions within a small radius around the current agent position and attempting to navigate toward them. 
Candidate positions are randomly generated until the agent resumes motion.
If the agent fails to recover after $N_{\text{rec}}$ attempts, the current target object is discarded and the agent proceeds to the next one. 
The numerical values of the parameters $\tau_s$, $\epsilon_p$, and $N_{\text{rec}}$ are reported in Table~\ref{tab:data_collection_config}.

\textbf{Experimental configuration parameters.}
\label{sec:params}
Tab. \ref{tab:data_collection_config} reports a summary of the parameters and values used for data collection.

\begin{table}[t!]
\centering
\small
\caption{Parameters used during trajectory collection. 
}
\label{tab:data_collection_config}
\begin{tabular}{lll}
\toprule
\textbf{Symbol} & \textbf{Description} & \textbf{Value} \\
\midrule
$\alpha$ & Weighting coefficient for viewpoint selection & 0.7 \\
$A_{\min}$ & Minimum area of high-disagreement regions & 100 pixels \\
$\mathcal{R}$ & Viewpoint sampling radii & $\{0.5,1.0,2.0\}$ m \\
$N_r$ & Candidate viewpoints per radius & 30 \\
$N_{\min}$ & Minimum viewpoints per target object & 5 \\
$N_{\max}$ & Maximum viewpoints per target object & 20 \\
$\tau_s$ & Sliding window for stuck detection & 5 steps \\
$\epsilon_p$ & Minimum displacement threshold & 0.15 m \\
$N_{\text{rec}}$ & Maximum recovery attempts & 5 \\
\bottomrule
\end{tabular}
\end{table}


\section{LoRA Fine-Tuning}
\label{sec:supp_impl}
To efficiently fine-tune the pretrained Vision–Language backbone, we adopt Low-Rank Adaptation (LoRA) \cite{hu2022lora} as a parameter-efficient training strategy. Rather than updating all model parameters, LoRA injects trainable low-rank matrices into selected linear layers, significantly reducing memory footprint and training cost while preserving performance.

We apply LoRA to both the attention projections and selected language modeling components of the VLM. The specific configuration used in all experiments is summarized in Table~\ref{tab:lora_config}.

\begin{table}[t!]
\centering
\small
\caption{LoRA configuration used for fine-tuning the Unified VLM.}
\label{tab:lora_config}
\begin{tabular}{ll}
\toprule
\textbf{Parameter} & \textbf{Value} \\
\midrule
Rank ($r$) & 16 \\
Scaling factor ($\alpha$) & 8 \\
Dropout & 0.05 \\
Target modules & 
\begin{tabular}[c]{@{}l@{}}
\texttt{q\_proj}, \texttt{k\_proj}, \texttt{v\_proj}, \texttt{o\_proj}, \\
\texttt{language\_model.embed\_tokens}, \texttt{lm\_head}
\end{tabular} \\
\bottomrule
\end{tabular}
\end{table}

This configuration allows the model to adapt both its multimodal attention mechanisms and its output language distribution, which is critical for jointly learning data association, object-centric captioning, and action prediction within a unified sequence modeling framework.

\section{Additional results}
\label{sec:additional_results}

\subsection{Memory Scalability}

\input{tex/memory_token}

To assess the scalability of the proposed memory representation, we quantify the evolution of memory token count throughout episodic exploration as novel objects are discovered and tracked. Fig.~\ref{fig:memory_objects_progression} shows the progression of memory tokens and the object detected during the exploration.
The mean memory token count exhibits a similar increasing trend compared to the number of actively tracked objects.
This indicates that tokenization scales predominantly with object cardinality, rather than trajectory duration. 
The absence of abrupt deviations from object count confirms that episodic memory remains structurally compact and organized during exploration. 
Furthermore, the number of memory tokens has low inter-episode variance, underscoring consistent performance across diverse scenes and trajectories.

These results provide empirical validation that our tokenization scheme yields a predictably bounded memory representation, in direct contrast to geometry-centric mapping approaches (e.g., dense point cloud accumulation), which inherently suffer from unbounded memory expansion.

\subsection{Qualitative results}

\begin{figure}[t!]
\centering
    \begin{subfigure}{0.35\textwidth}
        \includegraphics[width=\textwidth]{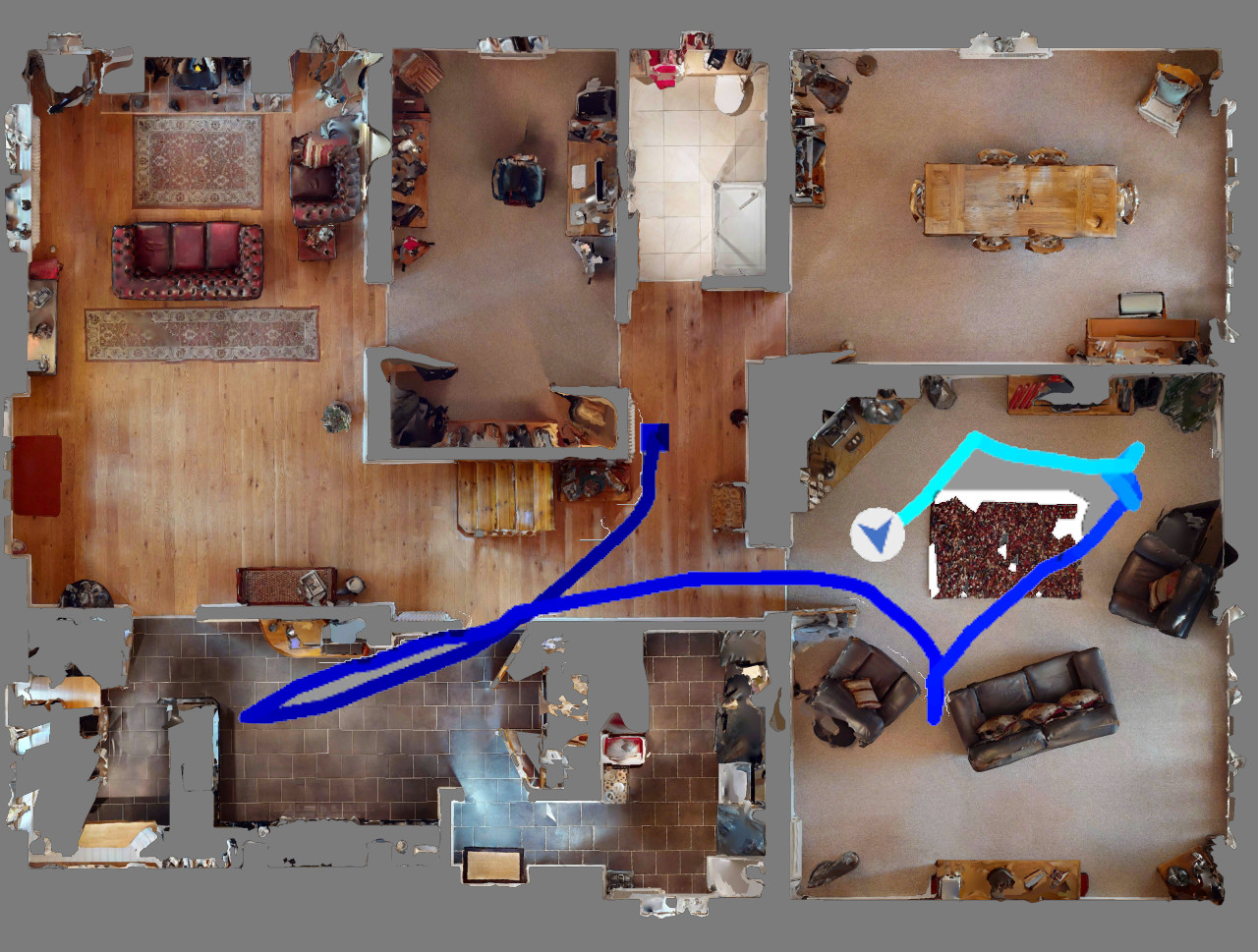}
        \caption{}
        \label{fig:first}
    \end{subfigure}
    \begin{subfigure}{0.55\textwidth}
        \includegraphics[width=\textwidth]{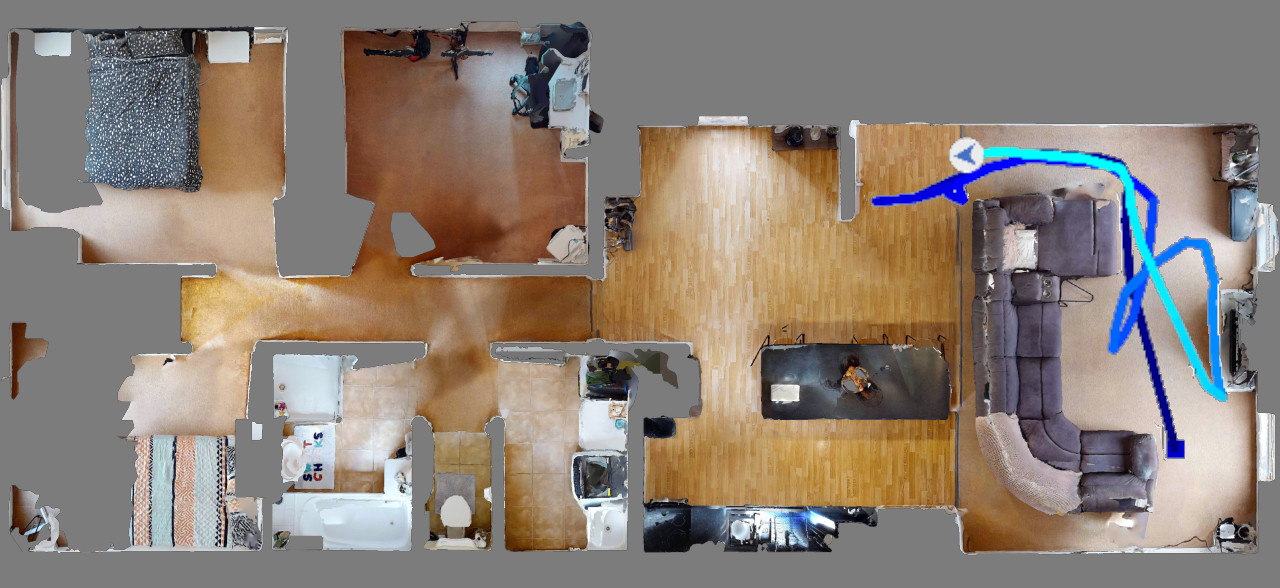}
        \caption{}
        \label{fig:first}
    \end{subfigure}
    \begin{subfigure}{0.3\textwidth}
        \includegraphics[width=\textwidth]{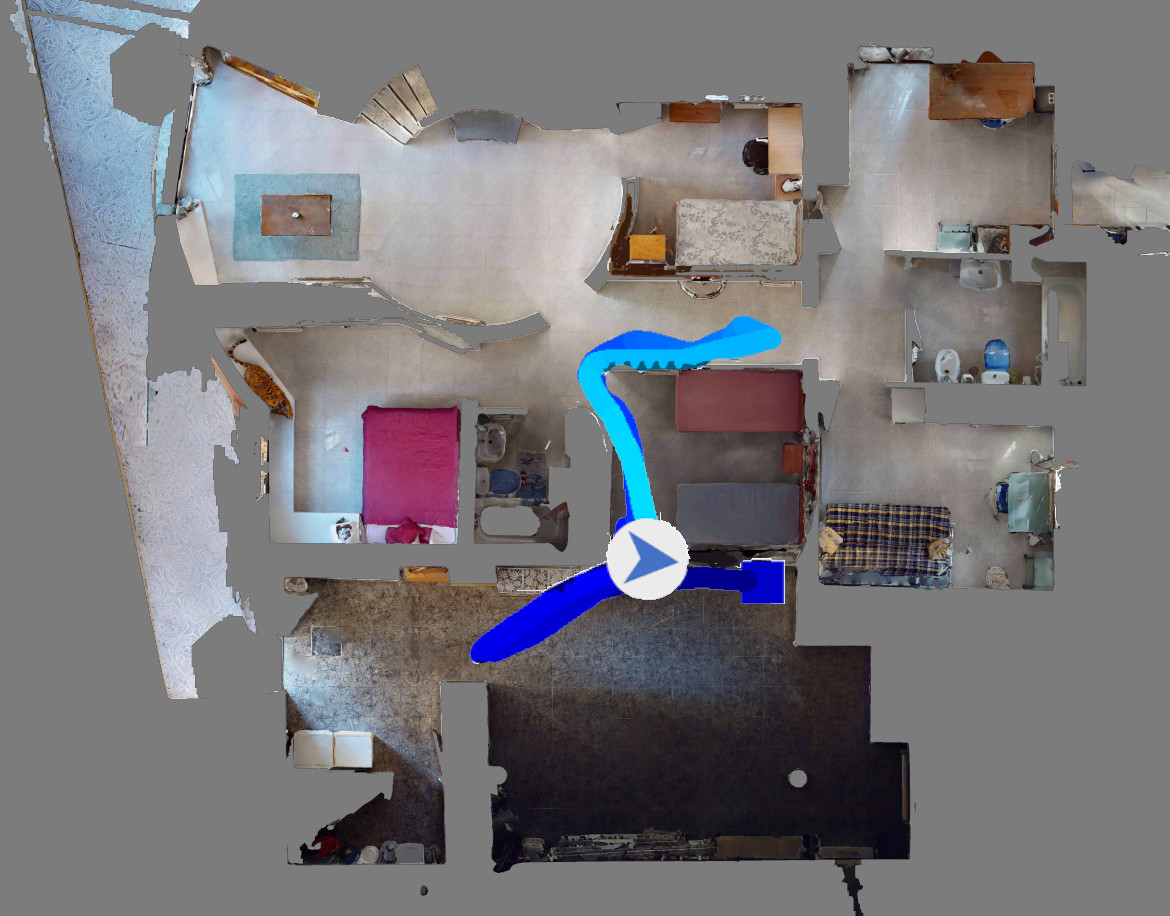}
        \caption{}
        \label{fig:first}
    \end{subfigure}
    \begin{subfigure}{0.45\textwidth}
        \includegraphics[width=\textwidth]{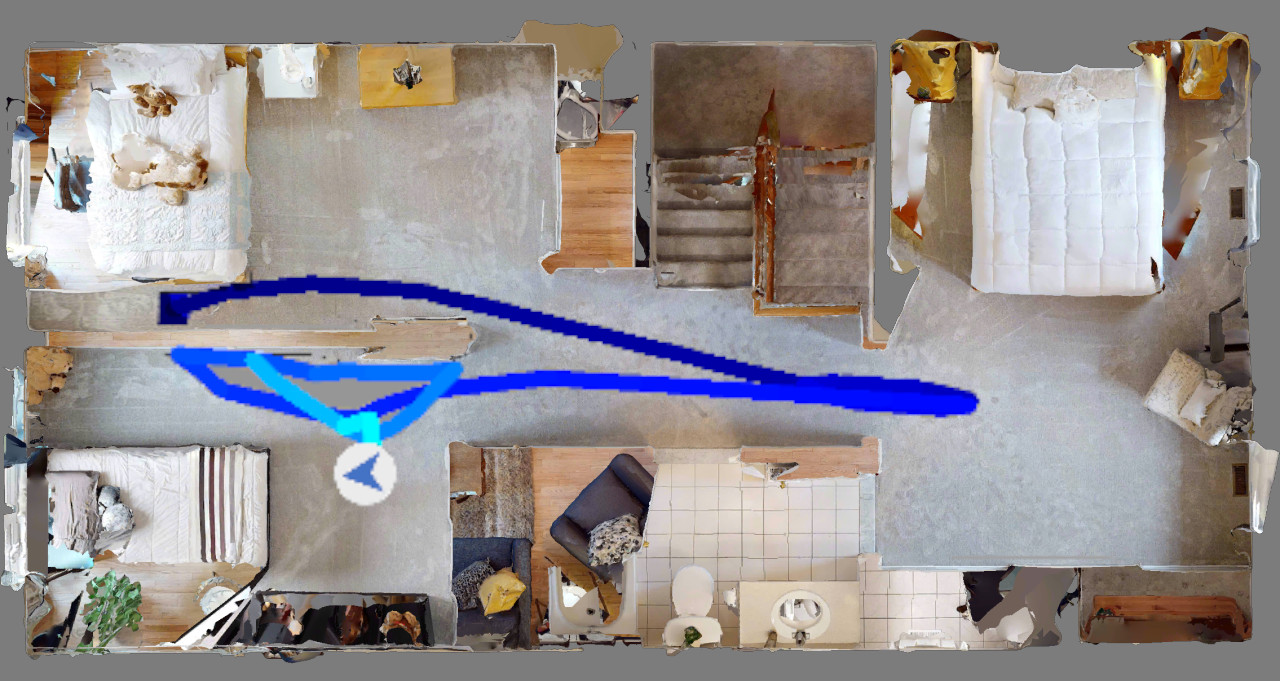}
        \caption{}
        \label{fig:first}
    \end{subfigure}    
    
  \caption{Examples of episodic trajectories on scenes from the HM3D test set. (a),(b),(c): the agent successfully revisits one or more objects from different viewpoints, while avoiding obstacles. (d): partial navigation failure where the agent hits a wall.}
  \label{fig:trajectories}
\end{figure}

Figure \ref{fig:trajectories} shows some qualitative examples of exploration trajectories on scenes from the HM3D test set.
Figures \ref{fig:trajectories}(a,b,c) depict a typical exploration episode in which the agent revisits one or more objects multiple times from different vantage points. The agent mostly focuses on one or two objects per episode due to the episode duration (400 steps).
In Figures \ref{fig:trajectories}d the agent, while still revisiting an object in the lower left corner of the map, collides with a nearby wall, therefore some of the steps are spent on recovering from the collision.

\begin{figure}[t!]
\centering
  \includegraphics[width=\textwidth]{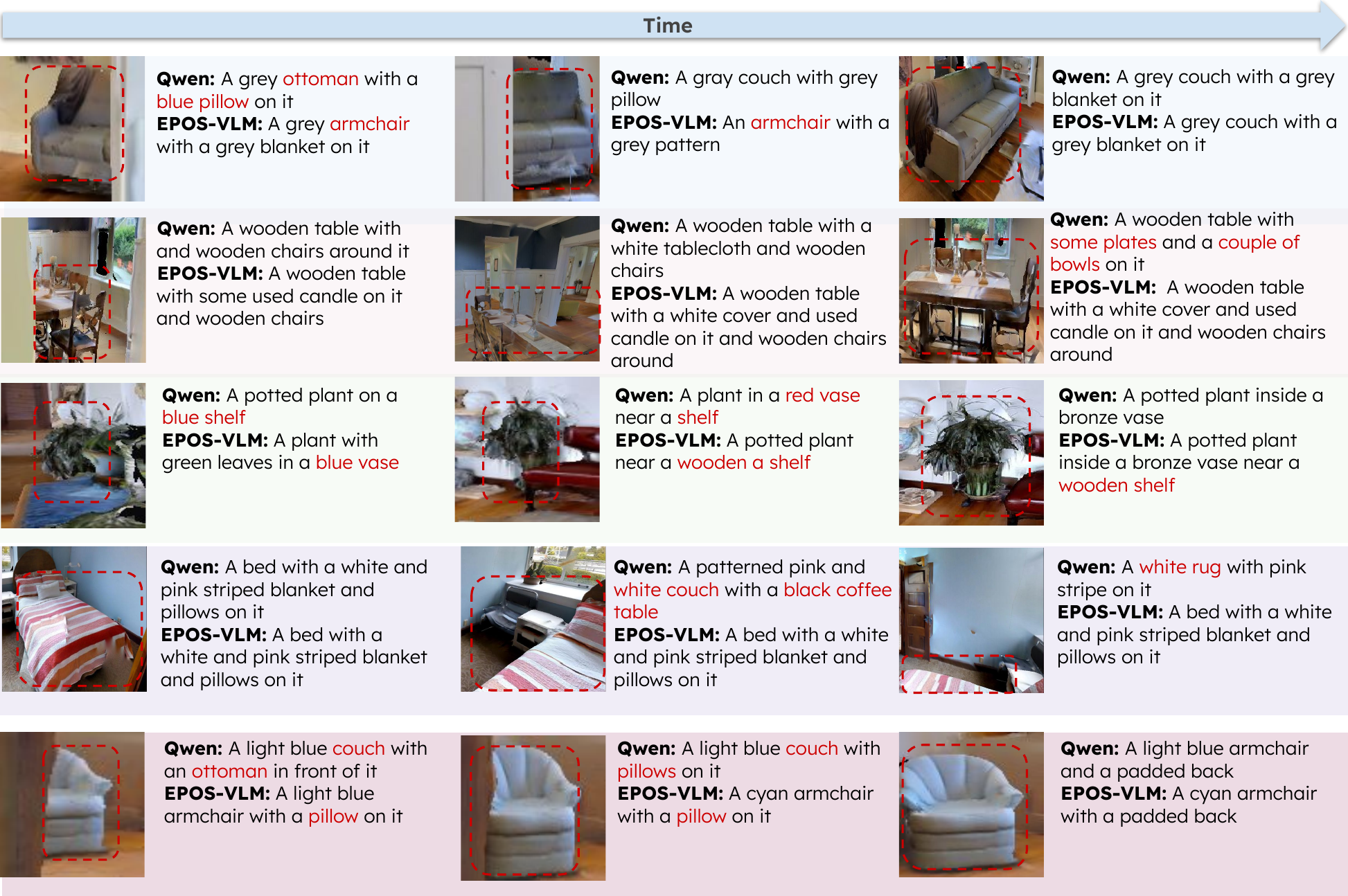}
  \caption{Additional qualitative examples for the vanilla and fine-tuned model of captions predicted over three views of the same object. Errors are highlighted in red.}
  \label{fig:captions}
\end{figure}

Figure~\ref{fig:captions} qualitatively compares randomly chosen object captions predicted by the vanilla Qwen3-VL model and the fine-tuned \ours\ across multiple views observed during exploration. 
The vanilla model predicts captions independently for each crop, often producing view-dependent descriptions or minor semantic variations for the same object. 
In contrast, \ours\ leverages episodic memory and explicit data association to preserve object identity across observations, resulting in more semantically consistent and accurate captions across viewpoints compared to Qwen3-VL-2B. 

\section{Additional pseudo-captioner comparisons}

To motivate the choice of Qwen3-VL-30B for the proposed 3D-CPS pseudo-captioner, we additionally evaluate it against two large vision-language models with similar number of parameters on the manually annotated HM3D test set (Tab.~\ref{tab:pseudocaption_hm3d}).
Overall, all models achieve strong results, indicating that the proposed pseudo-captioning strategy is valid across different VLM backbones. 
Among them, Qwen3-VL-30B achieves the best performance across all metrics, predicting pseudo-captions that are both more accurate with respect to human annotations and more semantically consistent. 
Based on these results, we adopt Qwen3-VL-30B as the backbone for 3D-CPS.
\input{tex/pseudocaption_vlm_comparrison}


%% file: tex/memory_token.tex
\pgfplotstableread{data/memory_tokens_stats.txt}\tokendata
\pgfplotstableread{data/objects_stats.txt}\objectdata

\begin{figure}[t!]
    \centering
    \begin{tikzpicture}
    \begin{axis}[
        width=0.48\linewidth,
        height=0.36\linewidth,
        xmin=-10, xmax=410,
        xtick={0,100,200,300,400},
        xlabel={Step number},
        ylabel={Memory tokens},
        label style={font=\footnotesize},
        tick label style={font=\footnotesize},
        grid=both,
        grid style={dashed, gray!20},
    ]

    \addplot[
        name path=lower,
        draw=none,
    ] table[
        x index=0,
        y expr=\thisrowno{1}-\thisrowno{2}
    ] {\tokendata};

    \addplot[
        name path=upper,
        draw=none,
    ] table[
        x index=0,
        y expr=\thisrowno{1}+\thisrowno{2}
    ] {\tokendata};

    \addplot[blue!30] fill between[of=lower and upper];

    \addplot[
        draw=blue,
        line width=1pt,
    ] table[
        x index=0,
        y index=1
    ] {\tokendata};

    \end{axis}
    \end{tikzpicture}
    \hfill
    \begin{tikzpicture}
    \begin{axis}[
        width=0.48\linewidth,
        height=0.36\linewidth,
        xmin=-10, xmax=410,
        xtick={0,100,200,300,400},
        xlabel={Step number},
        ylabel={Number of objects},
        label style={font=\footnotesize},
        tick label style={font=\footnotesize},
        grid=both,
        grid style={dashed, gray!20},
    ]

    \addplot[
        name path=lower,
        draw=none,
    ] table[
        x index=0,
        y expr=\thisrowno{1}-\thisrowno{2}
    ] {\objectdata};

    \addplot[
        name path=upper,
        draw=none,
    ] table[
        x index=0,
        y expr=\thisrowno{1}+\thisrowno{2}
    ] {\objectdata};

    \addplot[orange!35] fill between[of=lower and upper];

    \addplot[
        draw=orange!90!black,
        line width=1pt,
    ] table[
        x index=0,
        y index=1
    ] {\objectdata};

    \end{axis}
    \end{tikzpicture}
    \caption{
    Mean and standard deviation of memory token count (left) and number of objects (right) per step, averaged over all HM3D test episodes.}
    \label{fig:memory_objects_progression}
\end{figure}

%% file: tex/pseudocaption_vlm_comparrison.tex
\begin{table}[t!]
\centering
\caption{Pseudo-captioning performance on manually annotated HM3D~\cite{ramakrishnan2021habitatmatterport3ddatasethm3d} test set.
}
\label{tab:pseudocaption_hm3d}
\scriptsize
\begin{tabular}{lcccccc}
\toprule
\textbf{VLM} &\textbf{B4} & \textbf{M} & \textbf{RL} & \textbf{CI} & \textbf{SP} & \textbf{CS} \\
\midrule
InternVL3.5-30B~\cite{wang2025internvl35advancingopensourcemultimodal} & 
27.76 & 30.11 & 61.06 & 1.87 & 44.53 & 76.19 \\
Gemma3-27B~\cite{gemmateam2025gemma3technicalreport} & 
 26.54 & 29.12 & 58.95 & 1.68 & 42.06 & 73.36 \\
\rowcolor{lightgray}
\textbf{Qwen3-VL-30B}~\cite{yang2025qwen3technicalreport} & 
\textbf{28.54} & \textbf{31.97} & \textbf{61.13} & \textbf{1.92} & \textbf{44.73} & \textbf{78.26} \\
\bottomrule
\addlinespace[\belowrulesep]
\multicolumn{7}{l}{\parbox{0.7\linewidth}{\scriptsize{
KEYS -- $B4$: BLEU, $M$: METEOR, $RL$: ROUGE-L, $CI$: CIDEr, $SP$: SPICE, $CS$: cosine similarity between SBERT embedding of prediction and annotation.}}}
\end{tabular}
\end{table}